\documentclass[journal]{IEEEtran}

\usepackage{amsmath,amssymb,amsfonts}
\usepackage{graphicx}
\usepackage{url}
\usepackage{cite}
\usepackage{booktabs}
\usepackage{array}
\usepackage{stfloats}
\usepackage{tikz}
\usetikzlibrary{arrows.meta,positioning,calc,fit,backgrounds}
\usepackage{pgfplots}
\pgfplotsset{compat=1.18,compat/show suggested version=false}

\newcolumntype{P}[1]{>{\raggedright\arraybackslash}p{#1}}

\newcommand{\covY}{$\bullet$}   
\newcommand{\covP}{$\circ$}     
\newcommand{\covN}{--}          
\newcommand{\KB}{\mathcal{K}}
\newcommand{\semlev}[1]{\textbf{L#1}}

\begin{document}

\title{Semantic Communications for 6G:\\ From Deep Joint Source--Channel Coding to Foundation Models and Agentic AI}

\author{Yiheng~Zhang%
\thanks{E-mail: 2043749306@bupt.edu.cn.}%
\thanks{Manuscript prepared as a course survey assignment. The copyright of
this survey belongs to the author; the bibliography was assembled from
published survey reference lists and verified against the arXiv API.}%
}

\maketitle

\begin{abstract}
Semantic communication (SemCom) shifts the design objective of a communication
system from the faithful reproduction of bits to the faithful delivery of
\emph{meaning}, \emph{task-relevant information}, or \emph{actionable intent}.
Driven by the demands of sixth-generation (6G) networks---immersive extended
reality, autonomous machines, digital twins, and massive machine-type
intelligence---the field has grown from a handful of deep joint source--channel
coding (DeepJSCC) prototypes in 2019 into a broad research area spanning
semantic information theory, multimodal coding, knowledge-aware reasoning,
generative receivers, large AI models, multi-user networking, security, and
standardization. This survey provides a structured and critical account of that
evolution. We first formalize a unified three-level system model that maps
Weaver's technical, semantic, and effectiveness levels onto modern learning-based
architectures, and we organize the literature along six cross-cutting
dimensions. We then trace the development lineage from 2019 to 2026, review
semantic coding and modality-specific systems, and examine the two most
transformative recent threads: knowledge/reasoning-enabled SemCom and
generative/foundation-model-empowered SemCom. Network-level design,
security and privacy, evaluation, and standardization are treated as
first-class topics rather than appendices. The survey is built on 169
bibliographic records whose titles, authors, and dates were individually
verified; 142 of them (84.0\%) date from 2023 or later, including 74 records
from 2025--2026. We conclude with open problems in semantic information theory,
trustworthy generative reconstruction, knowledge-base synchronization,
coexistence with legacy bit-oriented infrastructure, and measurable
standardization.
\end{abstract}


\section{Introduction}
\IEEEPARstart{S}{ince} Shannon's 1948 formulation of the mathematical theory of
communication~\cite{Shannon}, the design of communication systems has been
dominated by a \emph{syntactic} objective: reproduce, at the receiver, the
sequence of symbols emitted by the source with arbitrarily small error. The
separation theorem, which permits independent optimization of source coding and
channel coding~\cite{cover1991elements}, made this objective tractable and
produced the highly successful digital communication stack in use today. Weaver,
however, already observed in 1953 that Shannon's theory addresses only the
\emph{technical} problem (Level~A) and deliberately abstracts away the
\emph{semantic} problem (Level~B, how precisely the transmitted symbols convey
the desired meaning) and the \emph{effectiveness} problem (Level~C, how
effectively the received meaning influences conduct)~\cite{Weaver}. For seven
decades these two upper levels remained outside engineering practice, largely
because ``meaning'' resisted formal measurement.

The 6G era has changed this. Immersive extended reality, holographic
telepresence, cooperative autonomous driving, industrial digital twins,
satellite--terrestrial integration, and swarms of embodied AI agents generate
data volumes that cannot be supported by brute-force increases in spectral
efficiency, while simultaneously requiring \emph{decisions} rather than exact
pixel or waveform reconstructions. Transmitting every bit is both unaffordable
and unnecessary: a surveillance camera needs to convey ``a pedestrian is
crossing,'' not the exact JPEG coefficients of the frame. Semantic
communication (SemCom) responds by transmitting only the information that is
relevant to the receiver's task, and by allowing the receiver to reconstruct
or reason about the rest using shared knowledge~\cite{SemanticSurvey4,ZLuSurvey,ChristinaLessdata}.

The modern, learning-based phase of the field began with DeepJSCC for wireless
image transmission~\cite{Bourtsoulatze2019DeepJSCC} and DeepSC for
text~\cite{HuiqiangText}, and rapidly diversified: task-oriented multi-user
systems~\cite{HuiqiangMultimodal,HuiqiangMultimodalVQA},
Transformer-based coding~\cite{YangkeImageWITT,DanlanImage}, robustness against
semantic noise~\cite{QiyuImageRobust,QiyuImage}, knowledge-graph-driven
reasoning~\cite{FuhuiReasoning2,BingyanReasoningKnowledge}, causal and
neuro-symbolic pragmatics~\cite{ThomasReason1,ThomasPragmatic}, generative
receivers~\cite{EcenazImage,TongAIGC,GenerativeAISem}, and large AI
models~\cite{GenerativeAI_Ref3,FeiboAIGCLarge,ZhenyiAIGC}. By 2025--2026 the
frontier had moved again, toward multimodal large language models (MLLMs),
direct LLM-to-LLM semantic exchange~\cite{arxiv251003215,arxiv251103945},
token-level communication~\cite{arxiv260910714,arxiv251101202}, agentic and
embodied communication~\cite{arxiv251223294,arxiv260935936,arxiv261029299},
and explicit standardization
roadmaps~\cite{arxiv250912758,arxiv260208675}. This acceleration is visible in
our own reference corpus: Fig.~\ref{fig:years} shows that 142 of the 169 works
cited here were published in 2023 or later.

\subsection{Why Another Survey?}
A substantial number of surveys already exist. Early broad treatments covered
fundamentals, applications, and challenges~\cite{SemanticSurvey4} and the
network/architecture perspective~\cite{SemanticSurvey0}. Specialized surveys
addressed metrics~\cite{SemanticSurvey6}, goal-oriented
communication~\cite{SemanticSurvey5}, resource management and
security~\cite{DWonSurvey}, secure SemCom~\cite{arxiv250100842},
intellicise/knowledge-centric networks~\cite{SemanticSurvey3,ChristinaLessdata},
space--air--ground--sea integration~\cite{MengSurvey}, generative
AI~\cite{ChengsiAIGC}, 6G techniques~\cite{WangSurvey}, tutorial-style
foundations~\cite{ZLuSurvey,SemanticSurvey1}, and, most recently, Theory of
Mind and DeepJSCC~\cite{arxiv250216468}, vision-centric
SemCom~\cite{arxiv260122202}, and robust
DeepJSCC~\cite{arxiv260404413}. Table~\ref{tab:surveys} compares their scope
with the present paper.

Two gaps motivate this survey. First, most existing surveys were written
before 2025 and therefore treat foundation models, LLM-to-LLM communication,
agentic semantics, and 6G standardization as outlook rather than as primary
material. Second, the literature is frequently organized either by
\emph{modality} (text/image/speech/video) or by \emph{technique}
(JSCC/knowledge/generative), which obscures the fact that the field is
progressing along Weaver's levels: from \emph{bit fidelity}, to
\emph{meaning fidelity}, to \emph{intent and effectiveness}. We organize the
paper around that progression while retaining a technology dimension, because
this makes both the historical lineage and the open theoretical problems
explicit.

\subsection{Scope and Literature-Selection Methodology}
\begin{figure}[!t]
\centering
\begin{tikzpicture}
\begin{axis}[
  width=0.97\columnwidth, height=4.8cm,
  ybar, bar width=6pt,
  xlabel={Publication year}, ylabel={Cited references},
  xlabel style={font=\scriptsize}, ylabel style={font=\scriptsize},
  xtick={2018,2019,2020,2021,2022,2023,2024,2025,2026},
  xticklabels={$\le$2018,2019,2020,2021,2022,2023,2024,2025,2026},
  xticklabel style={font=\tiny, rotate=35, anchor=east},
  ytick={0,10,20,30,40,50}, ymin=0, ymax=58,
  yticklabel style={font=\tiny},
  grid=major, grid style={dashed, gray!30},
  nodes near coords,
  nodes near coords align={vertical},
  every node near coord/.append style={font=\tiny, black!80},
  enlarge x limits=0.13,
  axis on top=false,
  bar shift=0pt
]
\addplot[fill=blue!45, draw=blue!70!black, line width=.4pt]
  coordinates {(2018,5)(2019,2)(2020,1)(2021,6)(2022,13)(2023,27)(2024,41)(2025,51)(2026,23)};
\end{axis}
\end{tikzpicture}
\caption{Year distribution of the 169 references cited in this survey,
computed directly from the bibliography. A total of 142 references (84.0\%)
are dated 2023 or later, and 74 (43.8\%) are dated 2025--2026, reflecting the
recent acceleration of the field and the recency emphasis of this review.}
\label{fig:years}
\end{figure}
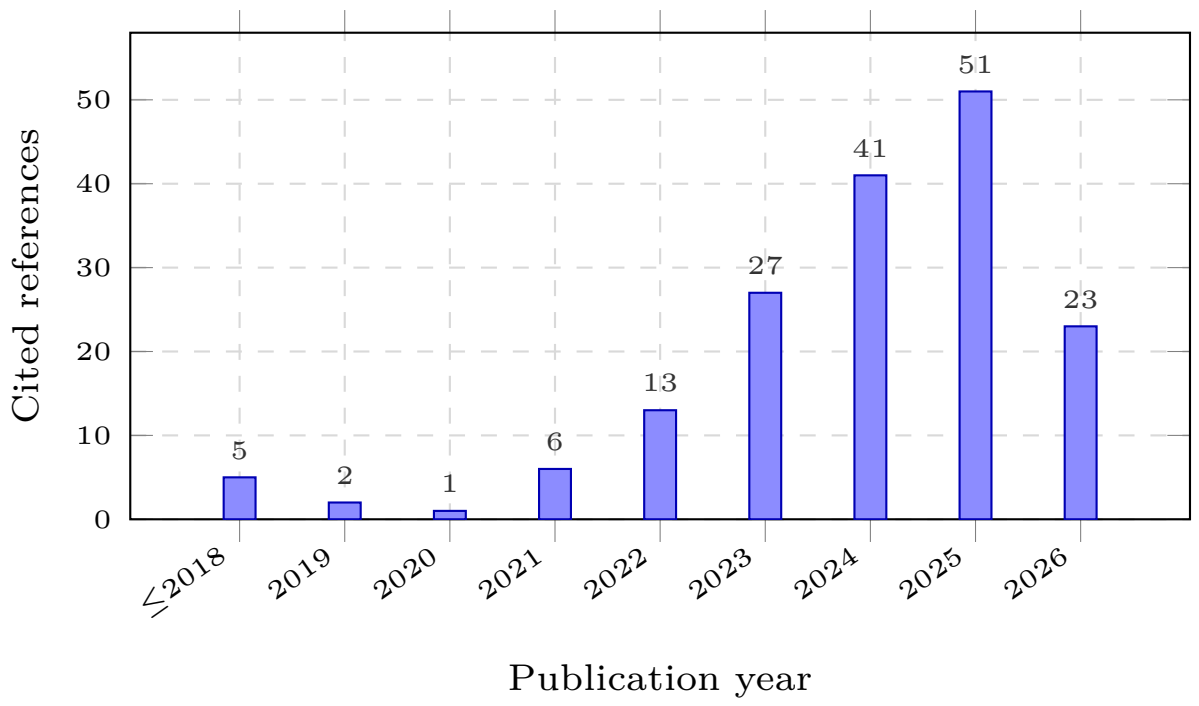

The reference corpus was assembled in three steps. (i) Published surveys with
extensive bibliographies were used as seed collections---in particular the
2025 IEEE Communications Surveys \& Tutorials article on Theory of Mind,
generative AI, and DeepJSCC~\cite{arxiv250216468}---from which peer-reviewed
journal and conference records were retained \emph{verbatim}. (ii) The arXiv
API was queried across eighteen topical search strings (surveys, task-oriented
communication, DeepJSCC, semantic information theory, knowledge graphs,
generative models, security, each modality, multi-user systems, and
standardization), yielding 774 unique records. (iii) Every preprint selected
for citation was re-queried by identifier to confirm its exact title, author
list, submission date, and, where available, journal reference; records whose
metadata did not match the collected title were discarded. The resulting 169
references contain 96 published archival items and 73 verified preprints,
spanning 1948--2026. No citation in this paper was generated from an
unverified source. Fig.~\ref{fig:years} reports the resulting year
distribution and is computed directly from the reference list.

\subsection{Contributions and Organization}
The contributions are fourfold:
\begin{enumerate}
\item A \emph{unified system model} that maps Weaver's three levels onto a
      learning-based encoder--channel--decoder pipeline with an explicit shared
      knowledge base, together with a compact set of optimization objectives
      (Section~\ref{sec:foundations}).
\item A \emph{development lineage and taxonomy} that connects the 2019--2022
      DeepJSCC wave, the 2022--2024 task/knowledge wave, and the 2024--2026
      generative, foundation-model, and agentic waves
      (Sections~\ref{sec:evolution}--\ref{sec:coding}).
\item A \emph{cross-cutting review} covering modality-specific systems,
      reasoning, generative AI, networking, security, evaluation, and
      standardization in a single consistent vocabulary, supported by six
      comparative tables and five original diagrams
      (Sections~\ref{sec:modality}--\ref{sec:evaluation}).
\item A \emph{critical agenda} of unresolved theoretical, measurement,
      trustworthiness, coexistence, and governance problems
      (Section~\ref{sec:challenges}).
\end{enumerate}
Section~\ref{sec:foundations} establishes foundations;
Section~\ref{sec:evolution} presents the taxonomy and timeline;
Section~\ref{sec:coding} reviews semantic coding;
Section~\ref{sec:modality} reviews modalities and tasks;
Section~\ref{sec:reasoning} reviews knowledge and reasoning;
Section~\ref{sec:generative} reviews generative and foundation-model systems;
Section~\ref{sec:network} reviews networks and emerging scenarios;
Section~\ref{sec:security} reviews security and privacy;
Section~\ref{sec:evaluation} reviews metrics, benchmarks and standardization;
Section~\ref{sec:challenges} lists open challenges; and
Section~\ref{sec:conclusion} concludes.

\section{Foundations and a Unified System Model}
\label{sec:foundations}

\subsection{From Bit Fidelity to Meaning and Effectiveness}
Let $\mathbf{s}\in\mathcal{S}$ denote a source signal (a sentence, an image, a
waveform, or a multimodal tuple) and let $t$ denote the receiver's
\emph{task}, e.g., classification, question answering, reconstruction, or
control. A conventional system implements a cascade of source coding and
channel coding; its design objective is
\begin{equation}
\min_{f,g}\;\mathbb{E}\!\left[d_{\mathrm{syn}}(\mathbf{s},\hat{\mathbf{s}})\right]
\quad \mathrm{s.t.}\quad \mathbb{E}\!\left[\|\mathbf{x}\|^2\right]\le P,\;\;
\mathrm{Pr}(\hat{\mathbf{s}}\neq\mathbf{s})\rightarrow 0,
\label{eq:shannon}
\end{equation}
where $f$ and $g$ are the source/channel encoder and decoder, $d_{\mathrm{syn}}$
is a symbol- or pixel-level distortion, and $P$ is a power constraint.
Equation~\eqref{eq:shannon} is indifferent to \emph{what the data are for}:
every bit is treated as equally valuable.

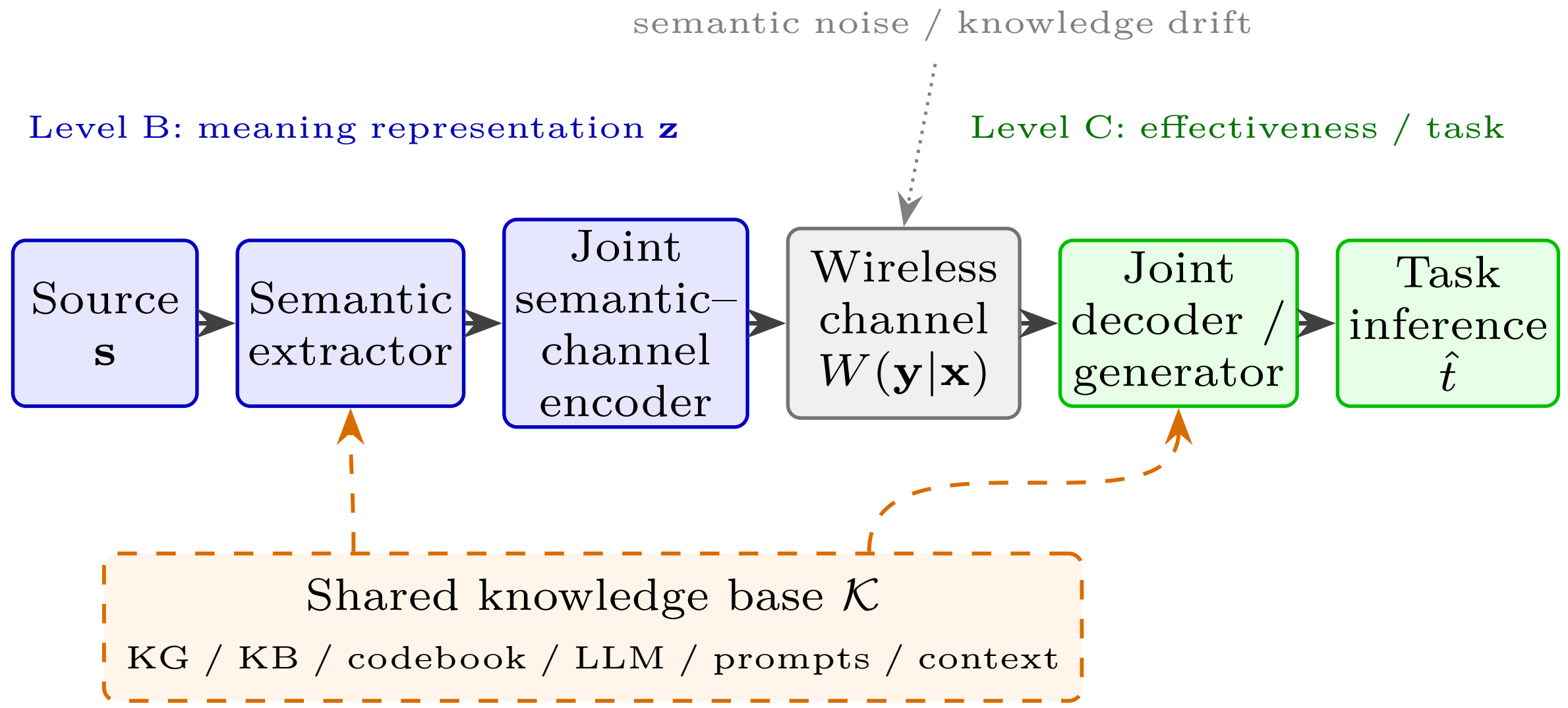
\begin{figure*}[!t]
\centering
\resizebox{0.86\textwidth}{!}{%
\begin{tikzpicture}[
  font=\scriptsize, >=Stealth,
  blk/.style={draw=#1!75!black, fill=#1!10, rounded corners=2pt,
              minimum width=10mm, minimum height=9mm, align=center,
              inner sep=1.6pt, line width=.55pt},
  arr/.style={->, line width=.7pt, black!75},
  dasharr/.style={->, dashed, line width=.6pt, orange!85!black}
]
\node[blk=blue] (src) {Source\\ $\mathbf{s}$};
\node[blk=blue, right=2.0mm of src] (ext) {Semantic\\ extractor};
\node[blk=blue, right=2.0mm of ext] (enc) {Joint\\ semantic--\\ channel\\ encoder};
\node[draw=black!55, fill=gray!12, right=2.0mm of enc, rounded corners=2pt,
      minimum width=12mm, minimum height=9mm, align=center, line width=.55pt]
      (ch) {Wireless\\ channel\\ $W(\mathbf{y}|\mathbf{x})$};
\node[blk=green, right=2.0mm of ch] (dec) {Joint\\ decoder /\\ generator};
\node[blk=green, right=2.0mm of dec] (task) {Task\\ inference\\ $\hat{t}$};

\foreach \a/\b in {src/ext,ext/enc,enc/ch,ch/dec,dec/task}
  \draw[arr] (\a)--(\b);

\node[draw=orange!85!black, fill=orange!8, dashed, rounded corners=2pt,
      align=center, minimum width=44mm, minimum height=8mm,
      line width=.6pt] (kb) at (2.65,-1.65)
      {Shared knowledge base $\KB$\\[1pt]
       \tiny KG / KB / codebook / LLM / prompts / context};
\draw[dasharr] ([xshift=-13mm]kb.north) to[out=90,in=-90] (ext.south);
\draw[dasharr] ([xshift=15mm]kb.north) to[out=90,in=-90] (dec.south);

\node[font=\tiny\itshape, text=blue!70!black] at (1.35,1.05)
      {Level B: meaning representation $\mathbf{z}$};
\node[font=\tiny\itshape, text=green!45!black] at (6.15,1.05)
      {Level C: effectiveness / task};
\node[font=\tiny\itshape, text=gray] (nz) at (4.55,1.62)
      {semantic noise / knowledge drift};
\draw[->, draw=gray, densely dotted, line width=.5pt] (nz) -- (ch.north);
\end{tikzpicture}}
\caption{Unified semantic communication system model. Solid arrows carry data;
dashed arrows carry shared knowledge. The transmitter and receiver are coupled
not only by the physical channel $W$ but also by a shared knowledge base
$\KB$; mismatch in $\KB$ is a source of \emph{semantic noise} that has no
counterpart in Shannon's Level-A model. The blue/green encoder--decoder pair is
typically optimized end-to-end by DeepJSCC-style methods, with the physical
channel treated as a differentiable, non-trainable layer.}
\label{fig:arch}
\end{figure*}

SemCom replaces reconstruction with a semantic representation
$\mathbf{z}$ and, increasingly, with a task outcome $\hat{t}$. A generic
learning-based pipeline is
\begin{align}
\mathbf{z}&=f_{\theta}(\mathbf{s};\KB), &
\mathbf{x}&=e_{\theta'}(\mathbf{z}), &
\mathbf{y}&=W(\mathbf{x};\mathbf{h})+\mathbf{n},\nonumber\\
\hat{\mathbf{z}}&=d_{\phi}(\mathbf{y};\KB), &
\hat{t}&=\psi(\hat{\mathbf{z}}),
\label{eq:system}
\end{align}
where $\KB$ is a \emph{shared knowledge base} (a knowledge graph, a corpus, a
codebook, a prompt library, or a foundation model), $W(\cdot)$ is the physical
channel with state $\mathbf{h}$, and $\mathbf{n}$ includes both channel noise
and \emph{semantic noise}---mismatch between the transmitter's and receiver's
interpretations~\cite{QiyuImageRobust,HuiqiangText}. Fig.~\ref{fig:arch}
depicts the pipeline.

The corresponding design objective is no longer~\eqref{eq:shannon} but a
\emph{goal-oriented} one:
\begin{equation}
\max_{\theta,\theta',\phi}\;
\mathbb{E}\!\left[U(\hat{t},t)\right]
-\lambda\,\mathbb{E}\!\left[c(\mathbf{x})\right]
\quad \mathrm{s.t.}\quad
\mathbb{E}\!\left[d_{\mathrm{sem}}(\mathbf{z},\hat{\mathbf{z}})\right]\le\epsilon,
\label{eq:goal}
\end{equation}
where $U$ is a task utility (accuracy, decision value, control reward),
$c(\mathbf{x})$ is the communication/computation cost, and $d_{\mathrm{sem}}$
is a semantic distortion that is generally \emph{not} a pixel-wise or
symbol-wise metric. When $U$ reduces to reconstruction fidelity and $\KB$ is
empty,~\eqref{eq:goal} collapses back to a rate--distortion problem; when $U$
is a decision reward and $\KB$ carries world knowledge, it becomes a
pragmatic communication problem~\cite{SemanticSurvey5,EmilioReasoning}.

A useful information-theoretic restatement is the information bottleneck
objective for task-oriented encoding~\cite{SongjieImageRobust,ChristinaLessdata}:
\begin{equation}
\min_{p(\mathbf{z}|\mathbf{s})}\; I(\mathbf{s};\mathbf{z})
-\beta\, I(\mathbf{z};t),
\label{eq:ib}
\end{equation}
which explicitly trades \emph{how much is transmitted} against \emph{how much
of the task is preserved}. The unresolved difficulty, discussed in
Section~\ref{sec:challenges}, is that neither $I(\mathbf{z};t)$ nor
$d_{\mathrm{sem}}$ has a universally accepted operational definition. Several
recent works attempt such definitions via semantic entropy and semantic
alphabets~\cite{arxiv240114160,arxiv251101202}, category-theoretic
formulations~\cite{arxiv250411334}, and mathematical theories of semantic
communication~\cite{arxiv240114160}.

\subsection{Measuring Meaning: A Four-Tier Metric Hierarchy}
Because~\eqref{eq:goal} replaces a well-defined distortion with $U$ and
$d_{\mathrm{sem}}$, evaluation is the field's most persistent methodological
problem. Getu \emph{et al.} provided an early systematic treatment of metrics
for semantic and goal-oriented communication~\cite{SemanticSurvey6}; since
then the metric landscape has expanded to include learned perceptual scores,
LLM-based similarity, and visually defined fidelity indices for generative
receivers~\cite{arxiv250510405}. Table~\ref{tab:metrics} organizes the metrics
into four tiers. A practical observation is that \emph{no single metric is
sufficient}: a generative receiver can score well on perceptual realism while
hallucinating task-critical entities, and a task-oriented encoder can achieve
high classification accuracy while being useless for human
inspection~\cite{arxiv250510405,arxiv260208801}. Recent benchmark efforts
therefore report bit-level, semantic-level, and task-level measures
jointly~\cite{arxiv260208675}.

\begin{table}[!t]
\renewcommand{\arraystretch}{1.12}
\caption{A Four-Tier Hierarchy of Semantic Communication Metrics}
\label{tab:metrics}
\centering
\footnotesize
\begin{tabular}{@{}P{1.6cm}P{2.1cm}P{1.7cm}P{1.9cm}@{}}
\toprule
\textbf{Tier} & \textbf{Representative measure} & \textbf{What it captures}
& \textbf{Typical use} \\
\midrule
Level A & BER / BLER, PSNR, SSIM, spectral efficiency
& Symbol and waveform fidelity; inherited from~\eqref{eq:shannon}
& Baseline and link budget \\
Level B & BLEU / sentence similarity, BERTScore, CLIP score,
semantic entropy, semantic similarity
& Whether the intended \emph{meaning} survives transmission
& Text, image--text, cross-modal~\cite{HuiqiangText,arxiv251101202} \\
Perceptual & LPIPS, FID, VFI, human opinion score
& Plausibility of \emph{generatively reconstructed} content
& Diffusion / GAN receivers~\cite{EcenazImage,arxiv250510405} \\
Level C & Task accuracy, mAP, VQA accuracy, decision success,
goal attainment, latency, energy, VoI/AoI
& Whether the receiver \emph{acts} correctly and timely
& Goal-oriented and agentic systems~\cite{HuiqiangMultimodal,SemanticSurvey5} \\
\bottomrule
\end{tabular}
\end{table}

\section{Evolution and Taxonomy of the Field}
\label{sec:evolution}

\subsection{A Developmental Lineage, 2019--2026}
\begin{figure*}[!t]
\centering
\resizebox{\textwidth}{!}{%
\begin{tikzpicture}[font=\scriptsize, >=Stealth,
  box/.style={draw=#1!75!black, fill=#1!8, rounded corners=2pt,
              text width=23mm, align=center, inner sep=2pt,
              minimum height=11mm, line width=.55pt},
  yr/.style={font=\scriptsize\bfseries}]
\draw[->, line width=1.1pt, gray!65] (0,0) -- (17.4,0);
\foreach \x/\lab in {0.4/2019,2.8/2020--21,5.2/2022,7.6/2023,10.0/2024,12.4/2025,14.8/2026}
  {\draw[line width=.8pt, gray!65] (\x,-0.12) -- (\x,0.12);
   \node[yr, anchor=north] at (\x,-0.16) {\lab};}
\node[anchor=west, font=\scriptsize\itshape, gray!70!black] at (16.1,0.12)
      {Wave I \quad Wave II \quad Wave III};

\node[box=blue, anchor=south] (a1) at (0.4,0.42)
  {DeepJSCC: learned end-to-end image coding~\cite{Bourtsoulatze2019DeepJSCC}};
\node[box=blue, anchor=north] (a2) at (2.8,-1.75)
  {DeepSC + lite IoT; OFDM / bandwidth-agile JSCC~\cite{HuiqiangText,MingyuImage}};
\node[box=green, anchor=south] (a3) at (5.2,0.42)
  {Task-oriented multi-user, DeepWiVe, NTSCC, DeepJSCC-Q~\cite{HuiqiangMultimodal,Tze-YangVideo,JinchengImage}};
\node[box=green, anchor=north] (a4) at (7.6,-1.75)
  {KG / causal / ToM reasoning; semantic-noise robustness; WITT~\cite{FuhuiReasoning2,ThomasReason1,YangkeImageWITT}};
\node[box=orange, anchor=south] (a5) at (10.0,0.42)
  {Generative + diffusion receivers; large AI models; COMST surveys~\cite{TongAIGC,FeiboAIGCLarge,ZLuSurvey}};
\node[box=orange, anchor=north] (a6) at (12.4,-1.75)
  {MLLM / MoE / RAG SemCom; agentic AI; 6G standardization roadmap~\cite{arxiv250216418,arxiv251223294,arxiv250912758}};
\node[box=purple, anchor=south] (a7) at (14.8,0.42)
  {Token communication, open benchmarks, embodied/agentic tutorials~\cite{arxiv260910714,arxiv260208675,arxiv260935936}};
\foreach \x/\n in {0.4/a1,2.8/a2,5.2/a3,7.6/a4,10.0/a5,12.4/a6,14.8/a7}
  \draw[densely dotted, gray!70, line width=.5pt] (\x,0.14) -- (\n);
\end{tikzpicture}}
\caption{Development lineage of semantic communication. Wave~I (2019--2022)
established learned joint source--channel coding and sentence-level semantics;
Wave~II (2022--2024) introduced task objectives, explicit knowledge, and
reasoning; Wave~III (2024--2026) introduced generative receivers, foundation
models, agents, and standardization. Representative works are cited at each
milestone.}
\label{fig:timeline}
\end{figure*}
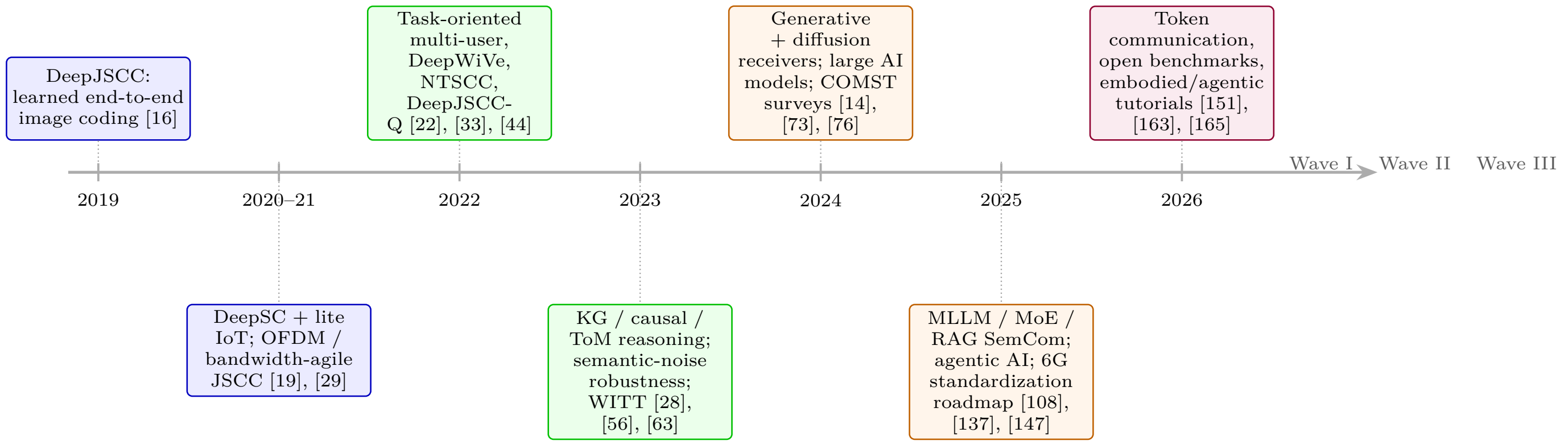

Fig.~\ref{fig:timeline} summarizes the field's development. Three waves can
be distinguished.

\emph{Wave I (2019--2022): learned end-to-end coding.} DeepJSCC showed that a
convolutional encoder--decoder trained over a differentiable channel model can
outperform separated source/channel coding at low signal-to-noise ratio (SNR)
and avoid the ``cliff effect''~\cite{Bourtsoulatze2019DeepJSCC}. This was
extended to OFDM and multipath fading~\cite{9500996,MingyuImage},
bandwidth-agile transmission~\cite{DavidImageBandwidth-agile}, constellation
constraints for digital modulation~\cite{Tze-YangImageDeepJSCC}, nonlinear
transform coding~\cite{JinchengImage}, and model-free training without an
explicit channel model~\cite{FaycalAitImage}. In parallel, DeepSC established
sentence-level semantic transmission with Transformer
backbones~\cite{HuiqiangText,HuiqiangText2,PeiwenText}. The important
conceptual step here was that \emph{the encoder is trained for a loss, not
designed for a code}; but the loss was still largely reconstruction-based, so
Wave~I sits mainly at Weaver's Level~B.

\emph{Wave II (2022--2024): tasks, knowledge, and reasoning.} Objective
functions moved from reconstruction to task success: multi-user visual
question answering~\cite{HuiqiangMultimodalVQA}, task-oriented image
transmission for edge inference~\cite{JiaweiImage,KangXuImage,JiawenImage}, and
unified multi-task systems~\cite{ZhangGuangyiMultimodal}. Knowledge graphs,
common languages, causal reasoning, neuro-symbolic AI, probabilistic logic, and
Theory of Mind were introduced to make the shared $\KB$ in~\eqref{eq:system}
explicit and
trainable~\cite{FuhuiReasoning2,BingyanReasoningKnowledge,Farshbafan,SeoProtocols,ThomasReason1,ThomasPragmatic,YongReasoning2,Frith}.
Robustness to semantic noise and adversarial mismatch became a distinct
subproblem~\cite{QiyuImageRobust,QiyuImage,SongjieImageRobust}.

\emph{Wave III (2024--2026): generative, foundation-model, and agentic
semantics.} Generative models changed the receiver from a decoder into a
\emph{synthesizer}: GAN- and diffusion-based JSCC reconstruct perceptually
plausible content from extremely compressed semantic
latents~\cite{EcenazImage,TongAIGC,EleonoraAIGC,GenerativeAISem}, and prompt- or
text-driven schemes achieve ultra-low bitrates by transmitting descriptions
rather than signals~\cite{PulkitAIGC,ZhengAIGCSpeech}. Large AI models and
LLMs then took over semantic extraction, importance assessment, knowledge
storage, and resource
control~\cite{GenerativeAI_Ref3,FeiboAIGCLarge2,FeiboAIGCLarge,ZhenyiAIGC,WantingAIGC,AIGCEdge2,ShuaishuaiAIGC}.
By 2025--2026, MLLMs, mixture-of-experts models, RAG, and direct LLM-to-LLM
exchange were standard
ingredients~\cite{arxiv250216418,arxiv250315722,arxiv250904801,arxiv251003215,arxiv251103945},
and the discussion shifted toward \emph{tokens} as the unit of transmission,
\emph{agents} as the communicating endpoints, and \emph{standardization} as an
engineering
requirement~\cite{arxiv251101202,arxiv260910714,arxiv251223294,arxiv260935936,arxiv250912758}.

\subsection{Proposed Taxonomy}
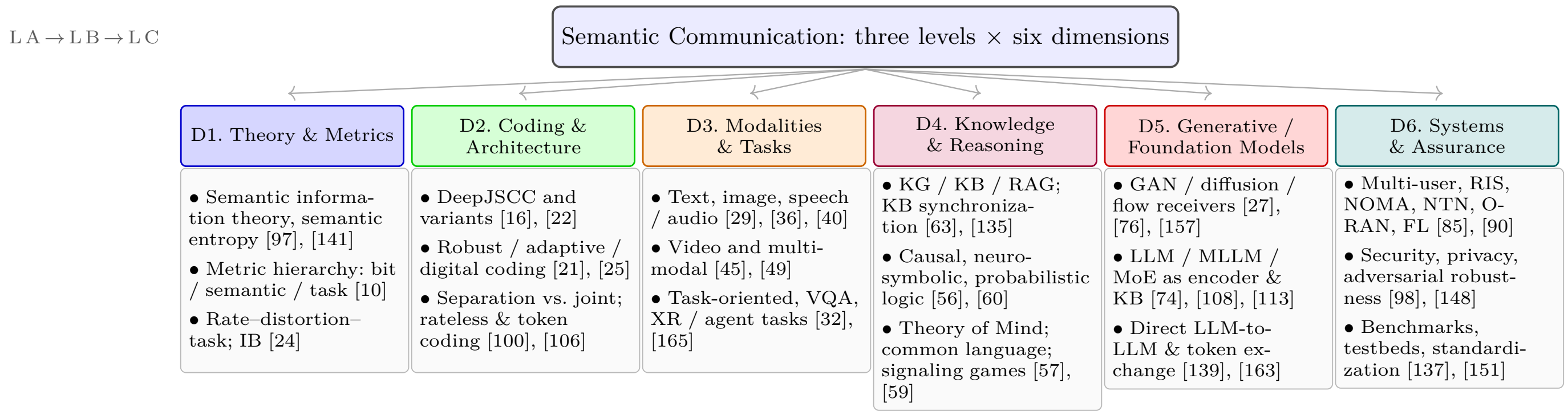
\begin{figure*}[!t]
\centering
\resizebox{\textwidth}{!}{%
\begin{tikzpicture}[font=\scriptsize,
  head/.style={draw=#1!80!black, fill=#1!16, rounded corners=2pt,
               text width=28mm, align=center, minimum height=8mm,
               inner sep=2pt, font=\scriptsize\bfseries, line width=.6pt},
  leaf/.style={draw=gray!55, fill=gray!4, rounded corners=2pt,
               text width=28mm, align=left, inner sep=3pt,
               minimum height=27mm, line width=.4pt},
  itm/.style={font=\tiny, inner sep=0pt}]
\node[draw=black!70, fill=blue!8, rounded corners=3pt, minimum width=48mm,
      minimum height=8mm, align=center, line width=.8pt,
      font=\small\bfseries] (root) at (9.05,2.05)
      {Semantic Communication: three levels $\times$ six dimensions};
\node[anchor=east, font=\scriptsize\itshape, text=gray!70!black] at (-0.15,2.05)
      {L\,A $\!\to\!$ L\,B $\!\to\!$ L\,C};

\node[head=blue, anchor=north west]   at (0,1.15)    {D1. Theory \& Metrics};
\node[head=green, anchor=north west]  at (3.05,1.15)  {D2. Coding \& Architecture};
\node[head=orange, anchor=north west] at (6.10,1.15)  {D3. Modalities \& Tasks};
\node[head=purple, anchor=north west] at (9.15,1.15)  {D4. Knowledge \& Reasoning};
\node[head=red, anchor=north west]    at (12.20,1.15) {D5. Generative / Foundation Models};
\node[head=teal, anchor=north west]   at (15.25,1.15) {D6. Systems \& Assurance};

\node[leaf, anchor=north west] at (0,0.32)
  {$\bullet$ Semantic information theory, semantic entropy~\cite{arxiv240114160,arxiv251101202}\\[3pt]
   $\bullet$ Metric hierarchy: bit / semantic / task~\cite{SemanticSurvey6}\\[3pt]
   $\bullet$ Rate--distortion--task; IB~\cite{SongjieImageRobust}};
\node[leaf, anchor=north west] at (3.05,0.32)
  {$\bullet$ DeepJSCC and variants~\cite{Bourtsoulatze2019DeepJSCC,JinchengImage}\\[3pt]
   $\bullet$ Robust / adaptive / digital coding~\cite{QiyuImage,Tze-YangImageDeepJSCC}\\[3pt]
   $\bullet$ Separation vs.\ joint; rateless \& token coding~\cite{arxiv250104285,arxiv250206095}};
\node[leaf, anchor=north west] at (6.10,0.32)
  {$\bullet$ Text, image, speech / audio~\cite{HuiqiangText,DanlanImage,ZhenziSpeech1}\\[3pt]
   $\bullet$ Video and multimodal~\cite{SixianVideo,ZhangGuangyiMultimodal}\\[3pt]
   $\bullet$ Task-oriented, VQA, XR / agent tasks~\cite{HuiqiangMultimodalVQA,arxiv260935936}};
\node[leaf, anchor=north west] at (9.15,0.32)
  {$\bullet$ KG / KB / RAG; KB synchronization~\cite{FuhuiReasoning2,arxiv250904801}\\[3pt]
   $\bullet$ Causal, neuro-symbolic, probabilistic logic~\cite{ThomasReason1,SeoProtocols}\\[3pt]
   $\bullet$ Theory of Mind; common language; signaling games~\cite{ThomasPragmatic,Farshbafan}};
\node[leaf, anchor=north west] at (12.20,0.32)
  {$\bullet$ GAN / diffusion / flow receivers~\cite{EcenazImage,TongAIGC,arxiv260724876}\\[3pt]
   $\bullet$ LLM / MLLM / MoE as encoder \& KB~\cite{ZhenyiAIGC,arxiv250216418,arxiv250315722}\\[3pt]
   $\bullet$ Direct LLM-to-LLM \& token exchange~\cite{arxiv251003215,arxiv260910714}};
\node[leaf, anchor=north west] at (15.25,0.32)
  {$\bullet$ Multi-user, RIS, NOMA, NTN, O-RAN, FL~\cite{WeizhiImageNon,Sun_S-RAN}\\[3pt]
   $\bullet$ Security, privacy, adversarial robustness~\cite{arxiv250100842,arxiv260101791}\\[3pt]
   $\bullet$ Benchmarks, testbeds, standardization~\cite{arxiv260208675,arxiv250912758}};

\foreach \x in {1.425,4.475,7.525,10.575,13.625,16.675}
  \draw[->, draw=gray!65, line width=.5pt] (9.05,1.62) -- (\x,1.30);
\end{tikzpicture}}
\caption{Proposed taxonomy. Six research dimensions (D1--D6) cut across
Weaver's three semantic levels; arrows from the root indicate that every
dimension contributes to all three levels, with recent work concentrating at
Level~C. Each leaf is anchored by representative verified references.}
\label{fig:taxonomy}
\end{figure*}

Fig.~\ref{fig:taxonomy} gives the taxonomy used throughout this survey. The
vertical axis follows Weaver's levels: \semlev{A} (bit/symbol fidelity),
\semlev{B} (meaning fidelity), and \semlev{C} (intent and effectiveness). The
horizontal axis follows six research dimensions: (D1) theory and metrics,
(D2) semantic coding and architecture, (D3) modalities and tasks,
(D4) knowledge and reasoning, (D5) generative AI and foundation models, and
(D6) systems---networking, security, and standardization. Unlike purely
modality-based taxonomies~\cite{arxiv260122202}, this arrangement makes it
explicit that a text system using DeepSC~\cite{HuiqiangText} and an LLM-based
image system~\cite{arxiv250112988} may sit at \emph{different semantic levels}
even though they share a coding backbone, and that a diffusion
receiver~\cite{TongAIGC} and a neuro-symbolic
reasoner~\cite{ThomasReason1} solve complementary problems.

\begin{table*}[!t]
\renewcommand{\arraystretch}{1.15}
\setlength{\tabcolsep}{3pt}
\caption{Comparison With Representative Existing Surveys
($\bullet$ = dedicated coverage, $\circ$ = partial/incidental coverage,
\texttt{--} = not covered)}
\label{tab:surveys}
\centering
\scriptsize
\begin{tabular}{@{}P{3.0cm}P{1.9cm}P{3.2cm}ccccccc@{}}
\toprule
\textbf{Survey} & \textbf{Venue / Year} & \textbf{Primary scope}
& \textbf{Th./Met.} & \textbf{JSCC} & \textbf{Modal.} & \textbf{Reason.}
& \textbf{GenAI} & \textbf{Net./Sec.} & \textbf{Eval./Std.} \\
\midrule
Yang \emph{et al.}~\cite{SemanticSurvey4} & IEEE COMST 2023 & Fundamentals, applications, challenges
& \covP & \covY & \covY & \covP & \covN & \covY & \covN \\
Lu \emph{et al.}~\cite{ZLuSurvey} & IEEE COMST 2024 & Tutorial-cum-survey
& \covY & \covY & \covY & \covP & \covP & \covY & \covN \\
Getu \emph{et al.}~\cite{SemanticSurvey6} & IEEE Access 2023 & Metrics for semantic/goal-oriented comm.
& \covY & \covP & \covP & \covP & \covN & \covP & \covP \\
Getu \emph{et al.}~\cite{SemanticSurvey5} & IEEE Access 2024 & Goal-oriented SemCom
& \covP & \covY & \covY & \covY & \covP & \covY & \covP \\
Zhang \emph{et al.}~\cite{SemanticSurvey3} & IEEE COMST 2024 & Intellicise wireless networks
& \covP & \covP & \covY & \covY & \covP & \covY & \covP \\
Chaccour \emph{et al.}~\cite{ChristinaLessdata} & IEEE COMST 2024 & Knowledge-centric, ``less data''
& \covY & \covP & \covP & \covY & \covP & \covY & \covP \\
Guo \emph{et al.}~\cite{SemanticSurvey0} & IEEE COMST 2024 & Architecture, security, privacy
& \covP & \covP & \covY & \covP & \covP & \covY & \covP \\
Won \emph{et al.}~\cite{DWonSurvey} & IEEE COMST 2024 & Resource, security, privacy
& \covP & \covP & \covP & \covN & \covN & \covY & \covN \\
Liang \emph{et al.}~\cite{ChengsiAIGC} & IEEE TCCN 2024 & Generative-AI-driven SemCom networks
& \covP & \covP & \covY & \covP & \covY & \covY & \covN \\
Meng \emph{et al.}~\cite{arxiv250100842} & arXiv 2025 & Secure SemCom
& \covP & \covP & \covY & \covP & \covP & \covY & \covN \\
Nguyen \emph{et al.}~\cite{arxiv250216468} & IEEE COMST 2025 & ToM, GenAI, DeepJSCC
& \covY & \covY & \covY & \covY & \covY & \covP & \covN \\
Wang \emph{et al.}~\cite{WangSurvey} & IEEE Access 2025 & 6G techniques and applications
& \covP & \covY & \covY & \covP & \covP & \covY & \covP \\
Cheng \emph{et al.}~\cite{arxiv260122202} & IEEE TNSE 2026 & Vision-centric SemCom
& \covP & \covY & \covY & \covP & \covY & \covP & \covP \\
Hong \emph{et al.}~\cite{arxiv260404413} & arXiv 2026 & Robust DeepJSCC
& \covP & \covY & \covP & \covN & \covP & \covP & \covP \\
\textbf{This survey} & 2026 & Levels A--C $\times$ D1--D6, 2019--2026
& \covY & \covY & \covY & \covY & \covY & \covY & \covY \\
\bottomrule
\end{tabular}
\end{table*}

\section{Semantic Coding and Deep Joint Source--Channel Coding}
\label{sec:coding}

\subsection{The DeepJSCC Family}
DeepJSCC maps an image directly to complex channel symbols through a
convolutional autoencoder trained over a differentiable channel
layer~\cite{Bourtsoulatze2019DeepJSCC}. Its two signature properties are the
absence of a ``cliff effect'' (graceful degradation as SNR falls) and superior
low-SNR/low-bandwidth robustness relative to separated compression plus
capacity-approaching channel codes. Subsequent work addressed practical
constraints: OFDM and multipath
fading~\cite{9500996,MingyuImage}, model-free training that does not require a
known channel gradient~\cite{FaycalAitImage}, variable bandwidth and adaptive
rate~\cite{DavidImageBandwidth-agile}, discrete constellations
compatible with digital modulation~\cite{Tze-YangImageDeepJSCC}, and nonlinear
transform coding that reintroduces an entropy model into the learned
coder~\cite{JinchengImage}. Transformer backbones, introduced
in~\cite{Vaswani,Dosovitskiy}, then entered the field: WITT~\cite{YangkeImageWITT}
and Swin-Transformer speech/video systems~\cite{ZhouSpeech} improved the
rate--perception tradeoff by exploiting SNR-adaptive attention, while the
information bottleneck formulation~\cite{SongjieImageRobust} and feature
compression for device--edge co-inference~\cite{JiaweiImage} connected DeepJSCC
to the broader edge-inference literature. A 2026 survey is devoted entirely to
robust DeepJSCC~\cite{arxiv260404413}, which indicates the maturity of this
subdirection.

\subsection{Robustness: Channel Noise versus Semantic Noise}
Two distinct impairments must be separated. \emph{Channel noise} is
additive/propagating distortion in $W(\cdot)$; \emph{semantic noise} arises
from distribution shift, mismatched training data, ambiguous symbols, or
inconsistent $\KB$ between the two endpoints~\cite{QiyuImageRobust,HuiqiangText}.
Masked VQ-VAE codebooks and adversarial training were proposed to make the
discrete semantic representation robust to both~\cite{QiyuImage}; robust
information bottleneck with digital modulation~\cite{SongjieImageRobust},
personality/saliency-aware transmission~\cite{JiawenImage}, and knowledge
distillation under imperfect channel state
information~\cite{arxiv250904005} refine the same idea. More recent work
examines \emph{where} in a Transformer to defend, using layer-wise adversarial
training~\cite{arxiv260913128}, formally verifies DNN-based SemCom against
generative adversarial noise~\cite{arxiv260208801}, and studies adversarially
robust image transmission~\cite{arxiv261008590}. Interference cancellation
diffusion models~\cite{arxiv250519983} and denoising diffusion
receivers~\cite{TongAIGC,arxiv250605710} treat the channel as a corruption
process to be inverted by a learned prior---a conceptual bridge to
Section~\ref{sec:generative}.

\subsection{Analog, Digital, and the Separation Debate}
A practical objection to DeepJSCC is that continuous-valued channel inputs are
incompatible with existing digital fronthaul, modulation, and retransmission
stacks. DeepJSCC-Q~\cite{Tze-YangImageDeepJSCC} and robust digital-modulation
IB~\cite{SongjieImageRobust} constrain outputs to finite constellations; HARQ
for sentence semantic transmission~\cite{PeiwenText} and semi-NOMA coexistence
of semantic and bit users~\cite{XidongText} address protocol-level
compatibility. The debate was reopened in 2025 by an LLM-based rethinking that
argues \emph{separate} source and channel coding remains preferable once
strong semantic compressors are
available~\cite{arxiv250104285}; a complementary proposal advocates rateless
JSCC as a blueprint for deployable 6G SemCom~\cite{arxiv250206095}. Meanwhile,
the discussion of what constitutes a ``semantic'' code led to calls for a
standardized coding workflow~\cite{arxiv250518637} and to token-oriented
formulations in which the transmitted unit is an LLM
token rather than a bit or a latent vector~\cite{arxiv251101202,arxiv260910714,arxiv260800368}.
We view this as an unresolved architectural question rather than a settled
result: it couples coding theory, hardware constraints, and the definition of
semantic distortion.

\section{Modality- and Task-Oriented Systems}
\label{sec:modality}
Table~\ref{tab:modality} summarizes representative systems by modality.

\subsection{Text}
DeepSC~\cite{HuiqiangText} introduced sentence-level semantic transmission
using a Transformer and a sentence-similarity loss, with a lite distributed
variant for IoT~\cite{HuiqiangText2}; DeepSC-S with HARQ added reliability
through incremental retransmission~\cite{PeiwenText}; resource allocation for
text SemCom~\cite{LeiTextresource} and heterogeneous semantic/bit
coexistence~\cite{XidongText} moved the text strand toward network
integration. Pre-trained language models were later used to weight semantic
importance at the word/sentence level~\cite{ShuaishuaiAIGC}, and LLMs now
recover punctured or corrupted text~\cite{arxiv250314831} or generate text
prompts as the transmitted representation~\cite{PulkitAIGC}.

\subsection{Image}
Image SemCom splits into reconstruction-oriented and task-oriented lines. The
former includes deep-learning-based semantic coding~\cite{DanlanImage},
wireless end-to-end image systems~\cite{LokumarambageImageWireless}, nonlinear
transform coding~\cite{JinchengImage}, and Transformers~\cite{YangkeImageWITT}.
The latter transmits only features needed for a downstream decision:
scene classification for
UAVs~\cite{KangXuImage}, personalized saliency~\cite{JiawenImage}, and
bottleneck features for device--edge
co-inference~\cite{JiaweiImage}. Generative JSCC~\cite{EcenazImage} and
semantic-slice-model propagation~\cite{SemanticSurvey3} occupy the middle
ground. Robustness to semantic noise~\cite{QiyuImageRobust,QiyuImage} and
knowledge-base-assisted CLIP retrieval~\cite{arxiv260405504} are recent
extensions; a 2026 vision-centric survey provides a complementary
treatment~\cite{arxiv260122202}.

\subsection{Speech and Audio}
DeepSC-S established semantic speech
transmission~\cite{ZhenziSpeech1}, followed by semantic-preserved
speech coding~\cite{TianxiaoSpeech}, streaming audio
(SemAudio~\cite{WeiSpeech}), Swin-Transformer speech
SemCom~\cite{ZhouSpeech}, and federated audio
SemCom~\cite{arxiv261128920}. Generative approaches now synthesize speech from
transmitted text or compact latents~\cite{ZhengAIGCSpeech,EleonoraAIGC}, which
is attractive for ultra-low-rate telephony but raises the fidelity/trustworthiness
question in Section~\ref{sec:security}.

\subsection{Video}
DeepWiVe~\cite{Tze-YangVideo} introduced learned wireless video delivery with
bandwidth allocation across frames; wireless deep video semantic
transmission~\cite{SixianVideo}, model-division video
SemCom~\cite{BaoVideo}, and VideoQA-SC~\cite{GuoVideo} extended it to
task-adaptive settings. Recent systems use latent diffusion for adaptive
streaming~\cite{arxiv250205695}, multimodal semantic fusion with large
models~\cite{arxiv250213838}, decoupled diffusion multi-frame
compensation~\cite{arxiv251102478}, and large-model-driven generative video
SemCom~\cite{arxiv250213838}. Talking-head video at ultra-low bitrate via
text~\cite{PulkitAIGC} is the most extreme instance of ``transmit meaning,
synthesize pixels.''

\subsection{Multimodal and Task-Oriented}
Cross-modal SemCom~\cite{LiAngMultimodal}, task-oriented multi-user VQA and
general multi-user task execution~\cite{HuiqiangMultimodalVQA,HuiqiangMultimodal},
a unified multi-task multimodal system~\cite{ZhangGuangyiMultimodal},
multimodal generative SemCom with latent diffusion~\cite{FuWeiqiMultimodal},
UAV-oriented virtual SemCom~\cite{RenChaoMultimodal}, and M4SC with
multimodal/multitask/multiuser support~\cite{arxiv250216418} illustrate the
convergence of modality strands. LLM-driven distributed integrated sensing and
SemCom~\cite{arxiv250518194}, MLLM-distilled coding~\cite{arxiv250907436}, and
multimodal LLM systems for immersive 6G~\cite{arxiv250704621} show that the
``modality'' boundary is increasingly defined by the foundation model's input
interface rather than by the physical signal.

\begin{table*}[!t]
\renewcommand{\arraystretch}{1.18}
\caption{Representative Modality- and Task-Oriented Semantic Communication Systems}
\label{tab:modality}
\centering
\footnotesize
\begin{tabular}{@{}P{1.9cm}P{4.8cm}P{3.8cm}P{3.0cm}P{2.9cm}@{}}
\toprule
\textbf{Modality / task} & \textbf{Representative systems}
& \textbf{Transmitted semantics} & \textbf{Dominant metrics}
& \textbf{Key references} \\
\midrule
Text & DeepSC, lite distributed SemCom, DeepSC-S+HARQ, semantic/bit semi-NOMA
& Sentence embeddings, word importance, LLM tokens
& Sentence similarity, BLEU, recovery accuracy
& \cite{HuiqiangText,HuiqiangText2,PeiwenText,LeiTextresource,XidongText,ShuaishuaiAIGC} \\
Image & DeepJSCC, WITT, NTSCC, semantic slice models, CLIP/KB-assisted coding
& Latent feature maps, saliency maps, codebook indices
& PSNR, MS-SSIM, LPIPS, classification accuracy
& \cite{Bourtsoulatze2019DeepJSCC,YangkeImageWITT,JinchengImage,DanlanImage,arxiv260405504} \\
Speech / audio & DeepSC-S, SemAudio, semantic-preserved speech, Swin-Transformer speech SemCom
& Spectrogram features, recognition transcripts, speaker/phoneme latents
& PESQ, STOI, WER, perceptual scores
& \cite{ZhenziSpeech1,WeiSpeech,TianxiaoSpeech,ZhouSpeech,EleonoraAIGC} \\
Video & DeepWiVe, wireless deep video SemCom, MDVSC, VideoQA-SC, diffusion-based streaming
& Frame latents, object--attribute--relation graphs, text prompts
& PSNR, VMAF, QA accuracy, latency
& \cite{Tze-YangVideo,SixianVideo,BaoVideo,GuoVideo,arxiv250205695,arxiv251102478} \\
Multimodal / task & VQA SemCom, cross-modal, unified multi-task, M4SC, UAV virtual SemCom
& Fused embeddings, task-specific bottlenecks, prompts
& Task accuracy, mAP, VQA score, energy
& \cite{HuiqiangMultimodalVQA,LiAngMultimodal,ZhangGuangyiMultimodal,RenChaoMultimodal,arxiv250216418} \\
Embodied / agent & Machine-language tokens, agent-native coding, interaction-aware SemCom
& Intent tokens, plans, beliefs, skill descriptors
& Task completion, latency, coordination reward
& \cite{arxiv250721454,arxiv260800368,arxiv261029299,arxiv261120399} \\
\bottomrule
\end{tabular}
\end{table*}

\section{Knowledge, Reasoning, and Pragmatic Semantics}
\label{sec:reasoning}

\subsection{Making the Shared Knowledge Base Explicit}
Equation~\eqref{eq:system} is only meaningful if $\KB$ is characterized.
Knowledge-graph-driven cognitive SemCom~\cite{FuhuiReasoning2} represents
background as triples and uses them to correct semantic errors at the receiver;
knowledge-enhanced receivers~\cite{BingyanReasoningKnowledge} and KG-based
multimodal representation fusion~\cite{SemanticSurvey3} extend this idea.
Correlated knowledge bases are modelled as a signaling
game~\cite{JinhoReasoningSemantic}, and language mismatch between users is
handled by a semantic channel equalizer~\cite{MohamedMultiuser}. A receiver-oriented
framework (ROSE)~\cite{ZhiyiOneReciver}, UAV KG-assisted object
detection~\cite{arxiv250203761}, explainable zero-shot KG
SemCom~\cite{arxiv250702291}, KG-RAG-assisted
SemCom~\cite{arxiv250904801}, and LLM-enabled knowledge
bases~\cite{arxiv260405504} progressively turn $\KB$ from a static graph into
a retrievable, updatable model component. Knowledge and power can be jointly
managed for secure operation~\cite{arxiv250415260}, and progressive SemCom with
knowledge-base reuse~\cite{arxiv260910112} addresses the storage cost of
maintaining $\KB$ at both ends.

\subsection{Reasoning, Pragmatics, and Theory of Mind}
Beyond shared facts, Level~C communication requires the receiver to infer
\emph{intent}. Curriculum learning with a common
language~\cite{Farshbafan}, probabilistic-logic
protocols~\cite{SeoProtocols}, contextual reasoning for semantics-native
communication~\cite{Hyowoon}, and reasoning ``over the
air''~\cite{YongReasoning2} treat meaning as something computed rather than
decoded. Neuro-symbolic causal reasoning combined with signaling
games~\cite{ThomasReason1}, causal SemCom for digital
twins~\cite{Thomasdigital}, and Theory-of-Mind-based pragmatic
communication~\cite{ThomasPragmatic,Frith} explicitly model the receiver's
belief state. The 6G-GOALS vision~\cite{EmilioReasoning} and the ``less data,
more knowledge'' perspective~\cite{ChristinaLessdata} position this as the
defining characteristic of AI-native 6G. Most recently, implicit-reasoning
elastic SemCom for AI-agent networks~\cite{arxiv261120399} and
interaction-aware agent collaboration~\cite{arxiv261029299} push pragmatics
toward multi-agent settings.

\subsection{Criticism}
The reasoning strand is conceptually the richest but empirically the least
comparable. Many systems are evaluated on bespoke simulation environments with
task-specific rewards, and the ``common language'' is often an emergent artifact
of joint training rather than an interpretable, standardized
representation~\cite{SemanticSurvey5,arxiv250216468}. Section~\ref{sec:evaluation}
returns to this reproducibility problem.

\section{Generative AI and Foundation-Model-Empowered Semantic Communication}
\label{sec:generative}

\subsection{Generative Receivers: From Decoding to Synthesis}
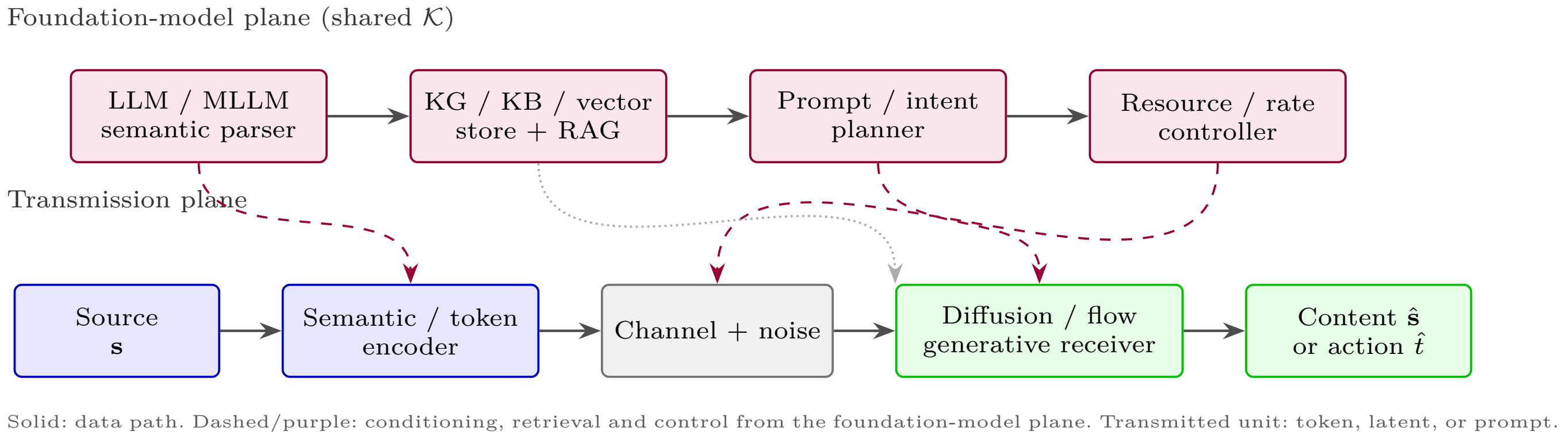
\begin{figure*}[!t]
\centering
\resizebox{\textwidth}{!}{%
\begin{tikzpicture}[font=\scriptsize, >=Stealth,
  blk/.style={draw=#1!75!black, fill=#1!10, rounded corners=2pt,
              minimum height=9mm, align=center, inner sep=2.2pt,
              line width=.6pt},
  arr/.style={->, line width=.7pt, black!70},
  sync/.style={->, dashed, line width=.6pt, purple!80!black}]
\node[font=\scriptsize\bfseries, text=black!80, anchor=west] at (0,2.2)
  {Foundation-model plane (shared $\KB$)};
\node[blk=purple, minimum width=25mm] (llm) at (2.0,1.25)
  {LLM / MLLM\\ semantic parser};
\node[blk=purple, minimum width=25mm, right=8mm of llm] (kg)
  {KG / KB / vector\\ store + RAG};
\node[blk=purple, minimum width=25mm, right=8mm of kg] (prompt)
  {Prompt / intent\\ planner};
\node[blk=purple, minimum width=25mm, right=8mm of prompt] (ctrl)
  {Resource / rate\\ controller};
\draw[arr] (llm)--(kg); \draw[arr] (kg)--(prompt); \draw[arr] (prompt)--(ctrl);

\node[font=\scriptsize\bfseries, text=black!80, anchor=west] at (0,0.42)
  {Transmission plane};
\node[blk=blue, minimum width=20mm] (src) at (1.2,-0.85) {Source\\ $\mathbf{s}$};
\node[blk=blue, minimum width=25mm, right=6mm of src] (tok)
  {Semantic / token\\ encoder};
\node[draw=black!55, fill=gray!12, rounded corners=2pt,
      minimum width=22mm, minimum height=9mm, align=center, line width=.6pt,
      right=6mm of tok] (ch) {Channel + noise};
\node[blk=green, minimum width=28mm, right=6mm of ch] (gen)
  {Diffusion / flow\\ generative receiver};
\node[blk=green, minimum width=22mm, right=6mm of gen] (out)
  {Content $\hat{\mathbf{s}}$\\ or action $\hat{t}$};
\foreach \a/\b in {src/tok,tok/ch,ch/gen,gen/out} \draw[arr] (\a)--(\b);

\draw[sync] (llm.south) to[out=-90,in=90] (tok.north);
\draw[sync] (prompt.south) to[out=-90,in=90] (gen.north);
\draw[sync] (ctrl.south) to[out=-90,in=90] (ch.north);
\draw[->, densely dotted, gray!65, line width=.6pt]
      (kg.south) to[out=-90,in=90] (gen.north west);

\node[font=\tiny, gray!70!black, anchor=west] at (0,-1.75)
  {Solid: data path. Dashed/purple: conditioning, retrieval and control from
   the foundation-model plane. Transmitted unit: token, latent, or prompt.};
\end{tikzpicture}}
\caption{Two-plane architecture of generative / foundation-model-empowered
SemCom. A foundation-model plane maintains the shared knowledge, prompts, and
control policy; a transmission plane carries a compact token or latent stream.
The receiver is a conditional generator rather than a decoder, so fidelity is
determined jointly by the transmitted semantics and by the quality and
synchronization of $\KB$.}
\label{fig:genpipeline}
\end{figure*}

The most consequential architectural change of Wave~III is that the receiver
may \emph{generate} the message from a very small semantic seed. Generative
JSCC~\cite{EcenazImage}, channel denoising diffusion models
(CDDM)~\cite{TongAIGC}, diffusion models for audio~\cite{EleonoraAIGC},
text-to-speech synthesis~\cite{ZhengAIGCSpeech}, and ultra-low-bitrate
talking-head video~\cite{PulkitAIGC} all exploit a strong prior to replace
missing signal. Cross-modal generative SemCom with joint semantic encoding and
prompt engineering~\cite{GenerativeAISem}, semantic feature decomposition with
large visual models~\cite{SenranAIGC}, and training-free diffusion
SemCom~\cite{arxiv250501209} reduce the training burden, while
semantics-guided diffusion JSCC~\cite{arxiv250101138}, latent-diffusion
denoising receivers~\cite{arxiv250605710}, interference-cancellation
diffusion~\cite{arxiv250519983}, and channel-realization flow
matching~\cite{arxiv260724876} target latency; generative-AI-based image
compression with edge optimization~\cite{arxiv250201675} addresses the
deployment side. Fig.~\ref{fig:genpipeline}
shows the resulting two-plane architecture, and Table~\ref{tab:generative}
lists representative systems. Dedicated 2025 surveys cover diffusion models
in the broader networking context~\cite{arxiv250801586,arxiv251108416}.

The price of this power is a new failure mode: a generative receiver can
produce content that is perceptually convincing but semantically
\emph{wrong}---hallucinating an object, a word, or a speaker identity that was
never transmitted. Visual fidelity indices with critical-information
embedding~\cite{arxiv250510405} and verification against adversarial
noise~\cite{arxiv260208801} are early responses; we return to the
trustworthiness question in Section~\ref{sec:security}.

\subsection{Large AI Models as Semantic Infrastructure}
Foundation models now occupy four roles: semantic encoder, knowledge store,
reasoner/controller, and interface to legacy networks. Design approaches and
open issues were articulated in~\cite{GenerativeAI_Ref3}; large AI
model-based~\cite{FeiboAIGCLarge2} and multimodal~\cite{FeiboAIGCLarge}
frameworks, LLM-enabled SemCom
systems~\cite{ZhenyiAIGC}, semantic importance-aware communications with
pre-trained LMs~\cite{ShuaishuaiAIGC}, RAG for GenAI-enabled
SemCom~\cite{AIGCEdge2}, and semantic-aware power
allocation~\cite{ChunmeiAIGC} illustrate the first three. MLLMs as universal
foundation models for AI-native wireless~\cite{ShengzheReasoning}, multi-user
generative SemCom with MLLMs~\cite{WantingAIGC}, AGI-driven generative
SemCom~\cite{arxiv250414947}, LLM-based image
transmission~\cite{arxiv250112988}, LLM-enabled data
transmission~\cite{arxiv250407431}, MoE-based zero-shot multi-task
SemCom~\cite{arxiv250315722}, MLLM-distilled coding~\cite{arxiv250907436}, and
generalizable foundation models for goal-oriented
communication~\cite{arxiv260907853} show the rapid broadening of this role.
AIGC-oriented frameworks for content
creation~\cite{GuangyuanAIGC}, metaverse integration~\cite{YijingAIGC}, and
secure resource allocation~\cite{ZhengAIGC} connect SemCom to application-layer
economics.

\subsection{LLM-to-LLM and Token Communication}
If both endpoints are themselves foundation models, the natural transmitted
unit is no longer a bit or a feature vector but a \emph{token} or a KV-cache
state. Cache-to-cache direct semantic communication~\cite{arxiv251003215} and
vector translation between LLMs~\cite{arxiv251103945} demonstrate this; a
semantic information theory ``from bits to tokens''~\cite{arxiv251101202}, the
semantic-to-token paradigm~\cite{arxiv260910714}, machine-language tokens for
task-oriented agent communication~\cite{arxiv250721454}, and agent-native joint
token compression coding and modulation~\cite{arxiv260800368} formalize it.
This is arguably the clearest Level~C instantiation: the ``meaning'' is an
internal model state that only another model can interpret. It also creates a
strong coupling to model versioning and alignment, since two differently
fine-tuned LLMs may not share $\KB$ even if they share vocabulary.

\begin{table*}[!t]
\renewcommand{\arraystretch}{1.16}
\caption{Representative Generative-AI and Foundation-Model-Empowered SemCom Systems}
\label{tab:generative}
\centering
\footnotesize
\begin{tabular}{@{}P{1.1cm}P{4.8cm}P{3.0cm}P{5.6cm}P{1.7cm}@{}}
\toprule
\textbf{Year} & \textbf{System / contribution} & \textbf{Model class}
& \textbf{Role of the generative model} & \textbf{Ref.} \\
\midrule
2023 & Txt2Vid: talking-head video via text & Text + GAN synthesis
& Replaces pixels by a textual description at ultra-low bitrate
& \cite{PulkitAIGC} \\
2023 & Generative JSCC for images & GAN / generative prior
& Reconstructs perceptually plausible images from compressed latents
& \cite{EcenazImage} \\
2023 & Semantic importance-aware communications & Pre-trained LM
& Weights words by semantic importance before transmission
& \cite{ShuaishuaiAIGC} \\
2024 & CDDM: channel denoising diffusion models & Diffusion
& Treats channel corruption as a reverse diffusion process
& \cite{TongAIGC} \\
2024 & Cross-modal generative SemCom + prompt engineering & Prompt + generative model
& Jointly optimizes semantic encoding and prompts for mobile AIGC
& \cite{GenerativeAISem} \\
2024 & Large AI model-based SemCom / foundation-model design & Large AI model
& Semantic extraction, KB, and resource orchestration
& \cite{FeiboAIGCLarge2,GenerativeAI_Ref3,ZhenyiAIGC} \\
2024 & RAG for GenAI-enabled SemCom & Retrieval-augmented generation
& Supplies up-to-date, receiver-aligned knowledge
& \cite{AIGCEdge2} \\
2025 & M4SC; MoE zero-shot multi-task; LLM image transmission & MLLM / MoE / LLM
& Unified multimodal, multitask, multiuser semantics
& \cite{arxiv250216418,arxiv250315722,arxiv250112988} \\
2025 & Training-free diffusion SemCom; LDM denoising receiver & Diffusion / LDM
& Removes the need for joint end-to-end retraining
& \cite{arxiv250501209,arxiv250605710} \\
2025 & AGI-driven generative SemCom; agentic-AI foundations & AGI / agent framework
& Positions SemCom inside an agent perception--action loop
& \cite{arxiv250414947,arxiv251223294} \\
2025 & Cache-to-cache; direct LLM-to-LLM vector translation & LLM internals
& Exchanges KV-cache states / latent vectors instead of text or bits
& \cite{arxiv251003215,arxiv251103945} \\
2026 & Channel-realization flow matching; generalizable foundation models & Flow / foundation model
& Low-latency generative reconstruction; cross-task generalization
& \cite{arxiv260724876,arxiv260907853} \\
2026 & From semantic to token communication & Token-level theory
& Re-defines the transmitted unit and the associated information measure
& \cite{arxiv260910714,arxiv251101202} \\
\bottomrule
\end{tabular}
\end{table*}

\section{Network-Level Design and Emerging Scenarios}
\label{sec:network}

\subsection{Multi-User Systems and Multiple Access}
Single-link SemCom does not scale directly. NOMA-enhanced multi-user
SemCom~\cite{WeizhiImageNon}, heterogeneous semantic/bit
semi-NOMA~\cite{XidongText}, task-oriented multi-user
designs~\cite{HuiqiangMultimodal,HuiqiangMultimodalVQA}, unified multi-task
multimodal systems~\cite{ZhangGuangyiMultimodal}, and MLLM-based multi-user
semantics~\cite{arxiv250216418} address asymmetry. Intent-aware
semantic-splitting multiple access~\cite{arxiv250701333} and coexistence,
multiple access, and satellite-network analyses~\cite{arxiv250611779} extend
this to 6G topologies; a 2026 survey treats non-terrestrial platform
constraints~\cite{arxiv260605216}. Semantic channel
equalization~\cite{MohamedMultiuser} and receiver-oriented
design~\cite{ZhiyiOneReciver} handle the fact that different users may hold
different $\KB$.

\subsection{Resource Allocation and Radio-Access Integration}
Because semantic importance is not uniform, resource allocation becomes
content-aware: text resource allocation~\cite{LeiTextresource},
DRL-driven dynamic allocation~\cite{HaijunImageDRL}, energy-efficient secure
allocation~\cite{ZhengAIGC}, semantic-aware power allocation with foundation
models~\cite{ChunmeiAIGC}, joint knowledge and power
management~\cite{arxiv250415260}, and content-aware adaptive-rate
coding~\cite{DavidImageBandwidth-agile,arxiv250216418}. Environment semantics
can aid mmWave beam and
blockage prediction~\cite{Yang_semantic}; RIS-aided joint active/passive
beamforming~\cite{Huang_semantic}, MIMO SemCom under imperfect
CSI~\cite{arxiv250904005}, SEM-O-RAN slicing~\cite{Puligheddu_SEM-O-RAN}, and
semantic-aware radio access networks~\cite{Sun_S-RAN} integrate SemCom into the
RAN. Federated training frameworks~\cite{Loc_SemanticFederated,Zhao_Semantic,arxiv260908593,arxiv261128920}
address the fact that $\KB$ and the encoder are often distributed across
devices that cannot share raw data.

\subsection{Application Scenarios}
IoT and lite SemCom~\cite{HuiqiangText2}, UAV
systems~\cite{KangXuImage,RenChaoMultimodal,arxiv250203761}, vehicular
networks and the Internet of Vehicles~\cite{arxiv250303767}, space--air--ground--sea
integration~\cite{MengSurvey}, 6G non-terrestrial
networks~\cite{NguyenOneTransmitterMultiple,arxiv260605216}, digital
twins~\cite{Thomasdigital}, metaverse and
AIGC~\cite{YijingAIGC,GuangyuanAIGC}, and quantum SemCom for
resource-efficient networking~\cite{Khalidquantum} show the breadth of
deployment contexts. The most recent scenario is \emph{embodied and agentic}:
wireless agentic AI with retrieval-augmented multimodal
perception~\cite{arxiv250523275}, long-horizon task
planning~\cite{arxiv260113685}, semantic Internet of Everything in the age of
agentic AI~\cite{arxiv260902924}, embodied SemCom for collective autonomous
agents~\cite{arxiv260935936}, and interaction-aware wireless agent
collaboration~\cite{arxiv261029299}. Goal-oriented communication for
multi-agent systems is now surveyed in its own right~\cite{arxiv250807720}.

\section{Security, Privacy, and Trustworthiness}
\label{sec:security}
SemCom changes the attack surface. Conventional encryption protects bits; a
SemCom system additionally exposes (i) a learned encoder whose latent space can
be inverted, (ii) a shared $\KB$ that may be poisoned or desynchronized, (iii)
a generative receiver that can be prompted to hallucinate, and (iv) an
agent-level intent channel that can be manipulated. Table~\ref{tab:security}
classifies the threat--defense landscape.

Early surveys framed the area around architecture, resource management,
security, and privacy~\cite{SemanticSurvey0,DWonSurvey}, followed by a dedicated
secure-SemCom survey~\cite{arxiv250100842}. Specific mechanisms include
blockchain-aided secure SemCom for
AIGC~\cite{Yijing2AIGC}, joint knowledge and power
management~\cite{arxiv250415260}, cross-layer security metrics and
optimization~\cite{arxiv250312818}, differential privacy over wiretap
channels~\cite{arxiv250418581,arxiv251023274}, diffusion-enabled secure
transmission against
eavesdropping~\cite{arxiv250505018,arxiv251101466},
privacy-preserving DeepJSCC with in-loop concept
erasure~\cite{arxiv260913393}, and KG-based secure
systems~\cite{arxiv250415260}. Robustness against adversarial semantic noise
is studied
in~\cite{QiyuImageRobust,arxiv260208801,arxiv260913128,arxiv261008590}.

The 2025--2026 literature identifies qualitatively new
threats. Latent manipulation is argued to be a distinct attack class in which
a small perturbation of the semantic latent changes the receiver's generated
\emph{meaning} without noticeably degrading link
quality~\cite{arxiv251203361}; generative and agentic AI are argued to require
a rethinking of the entire secure-SemCom threat
model~\cite{arxiv260101791}; and ``agentic eavesdropping,'' where an adversary
reconstructs intent by reasoning over intercepted semantic
tokens~\cite{arxiv260502204}, extends eavesdropping from content to intent.
We consider trustworthy generative reconstruction---including hallucination,
semantic leakage, and verifiability---the least mature part of the field.

\begin{table}[!t]
\renewcommand{\arraystretch}{1.13}
\caption{Threat Taxonomy and Representative Defenses in Semantic Communication}
\label{tab:security}
\centering
\footnotesize
\begin{tabular}{@{}P{1.6cm}P{2.45cm}P{2.35cm}P{1.1cm}@{}}
\toprule
\textbf{Threat} & \textbf{Mechanism} & \textbf{Representative defense}
& \textbf{Ref.} \\
\midrule
Semantic eavesdropping & Latent inversion reveals meaning without full decoding
& Wiretap-channel coding, jamming, physical-layer security
& \cite{DWonSurvey,arxiv250505018,arxiv250312818} \\
Differential privacy leakage & Reconstruction of private attributes from features
& Learnable DP, in-loop concept erasure
& \cite{arxiv250418581,arxiv251023274,arxiv260913393} \\
Adversarial / semantic noise & Perturbation of input or latent causes misreading
& Adversarial training, masked VQ-VAE, formal verification
& \cite{QiyuImage,QiyuImageRobust,arxiv260208801,arxiv260913128} \\
Latent manipulation & Small latent change alters generated meaning
& Cross-layer metrics, diffusion-aided secure JSCC
& \cite{arxiv251203361,arxiv250312818,arxiv251101466} \\
Knowledge poisoning / drift & Divergent or corrupted $\KB$ at the two ends
& KG-based security, joint knowledge--power management, KB reuse
& \cite{arxiv250415260,arxiv260910112,arxiv260502204} \\
Agentic eavesdropping & Intent inference from intercepted tokens
& Rethinking threat models for generative/agentic AI
& \cite{arxiv260101791,arxiv260502204} \\
Content forgery / hallucination & Generative receiver invents untransmitted facts
& Visual fidelity index, robust verification, adversarial training
& \cite{arxiv250510405,arxiv260208801,arxiv261008590} \\
\bottomrule
\end{tabular}
\end{table}

\section{Evaluation, Benchmarks, and Standardization}
\label{sec:evaluation}

\subsection{Metrics, Datasets, and Reproducibility}
Table~\ref{tab:metrics} gives the metric hierarchy; the practical difficulty
is that results reported in LPIPS, sentence similarity, mAP, and task success
cannot be compared across
papers~\cite{SemanticSurvey6,arxiv250510405}. Commonly used corpora include
natural-image sets for DeepJSCC evaluation, question-answering and detection
datasets for task-oriented systems, and speech/video corpora for streaming
studies; however, dataset splits, SNR definitions, channel models, and
bandwidth normalization vary widely. A critical 2025 position paper argues
that ``neural coding is not always semantic'' and calls for a standardized
coding workflow that separates learned compression from genuine semantic
abstraction~\cite{arxiv250518637}. This is, in our view, a prerequisite for
meaningful comparison: without an operational definition of semantic
distortion, the rate--semantic-distortion tradeoff cannot be benchmarked.

\subsection{Testbeds and Benchmarks}
Concrete evaluation infrastructure is emerging. GenSC-6G integrates
generative AI, quantum components, and SemCom in a prototype
testbed~\cite{arxiv250109918}; 6G-Bench provides an open benchmark for SemCom
and network-level reasoning with foundation models in AI-native
6G~\cite{arxiv260208675}; and standardized-coding, robust-DeepJSCC, and
vision-centric surveys consolidate evaluation
practice~\cite{arxiv260404413,arxiv260122202,arxiv250518637}. Testbeds remain
rare relative to simulation-only studies, which is a notable weakness.

\subsection{Standardization and the Path to Native AI}
Standardization is moving from vision statements to explicit roadmaps. A 2025
roadmap for native AI in 6G standardization identifies semantic communication
as a candidate
work item~\cite{arxiv250912758}, while the separation-vs-joint
debate~\cite{arxiv250104285,arxiv250206095}, the coding-workflow
proposal~\cite{arxiv250518637}, and the semantic-to-token
paradigm~\cite{arxiv260910714} delineate the technical choices that standards
must settle: unit of transmission, interface between semantic and bit layers,
knowledge-base synchronization protocol, and conformance
metrics. Coexistence with legacy infrastructure is addressed by semi-NOMA and
heterogeneous semantic/bit designs~\cite{XidongText,arxiv250611779}.
Table~\ref{tab:evalstd} lists the main evaluation and standardization assets.

\begin{table}[!t]
\renewcommand{\arraystretch}{1.14}
\caption{Evaluation Assets, Testbeds, and Standardization Activities}
\label{tab:evalstd}
\centering
\footnotesize
\begin{tabular}{@{}P{2.0cm}P{2.3cm}P{0.8cm}P{2.35cm}@{}}
\toprule
\textbf{Type} & \textbf{Item} & \textbf{Year} & \textbf{Ref.} \\
\midrule
Theory & Mathematical theory of SemCom; category-theoretic framework; bits-to-tokens
& 2024--25 & \cite{arxiv240114160,arxiv250411334,arxiv251101202} \\
Metrics & Systematic metrics survey; visual fidelity index
& 2023--25 & \cite{SemanticSurvey6,arxiv250510405} \\
Critique & ``Neural coding is not always semantic''; separation rethinking
& 2025 & \cite{arxiv250518637,arxiv250104285} \\
Benchmark & 6G-Bench: open SemCom + network reasoning benchmark
& 2026 & \cite{arxiv260208675} \\
Testbed & GenSC-6G prototype (GenAI + quantum + SemCom)
& 2025 & \cite{arxiv250109918} \\
Architecture & Rateless JSCC blueprint for 6G; token communication
& 2025--26 & \cite{arxiv250206095,arxiv260910714} \\
Roadmap & Native AI in 6G standardization
& 2025 & \cite{arxiv250912758} \\
Consolidation & Robust-DeepJSCC, vision, NTN, diffusion, multi-agent surveys
& 2025--26 & \cite{arxiv260404413,arxiv260122202,arxiv260605216,arxiv250801586,arxiv250807720} \\
\bottomrule
\end{tabular}
\end{table}

\section{Open Challenges and Future Directions}
\label{sec:challenges}
\begin{enumerate}
\item \emph{An operational semantic information theory.} Shannon's
\emph{separation} and \emph{channel capacity} have no generally accepted
semantic counterparts. Recent attempts---semantic entropy and semantic
alphabets~\cite{arxiv240114160,arxiv251101202}, category-theoretic
frameworks~\cite{arxiv250411334}, information bottleneck~\cite{SongjieImageRobust},
and knowledge-centric ``less data'' views~\cite{ChristinaLessdata}---are
promising but not yet unified. A useful theory should yield a computable
$d_{\mathrm{sem}}$, a rate--distortion--task function, and finite-blocklength
bounds that recover Shannon's results as a special case.
\item \emph{Comparable metrics and reproducible benchmarks.} Metric
proliferation (Table~\ref{tab:metrics}) makes cross-paper comparison
unreliable. Open benchmarks~\cite{arxiv260208675}, standardized coding
workflows~\cite{arxiv250518637}, and testbeds~\cite{arxiv250109918} should be
extended with common channel models, SNR/bandwidth normalization, dataset
splits, and reported confidence intervals.
\item \emph{Trustworthy generative reconstruction.} Hallucination, semantic
leakage, and verifiability of diffusion/LLM receivers are largely
unsolved~\cite{arxiv250510405,arxiv251203361,arxiv260208801}. A generative
receiver should be able to (a) flag content that is not supported by the
received semantics, and (b) be certified against bounded latent perturbations.
\item \emph{Knowledge-base lifecycle.} $\KB$ in~\eqref{eq:system} is treated as
static in most papers. Versioning, synchronization, drift detection,
personalization, and privacy-preserving updates across
heterogeneous users~\cite{MohamedMultiuser,arxiv260405504,arxiv260910112,arxiv260908593}
remain open; federated and RAG-based approaches are partial answers.
\item \emph{Coexistence and deployability.} Digital compatibility, HARQ,
constellation constraints, and hybrid semantic/bit
links~\cite{Tze-YangImageDeepJSCC,PeiwenText,XidongText,arxiv250104285,arxiv250206095}
must be reconciled with existing 3GPP stacks. Energy and latency costs of
running foundation models at the edge are rarely reported against the
transmission savings they claim.
\item \emph{Agentic semantics and governance.} When endpoints are autonomous
agents, the transmitted object is intent, and the threat model includes
deception, collusion, and intent
inference~\cite{arxiv260101791,arxiv260502204,arxiv261029299,arxiv260935936}.
Accountability, auditability, and alignment of communication policies will
require joint work with multi-agent safety rather than only with physical-layer
security.
\item \emph{From Level B to Level C evaluation.} Most ``goal-oriented''
systems still report static task accuracy on a fixed dataset. Effectiveness, in
Weaver's sense, is a closed-loop, temporal property: whether the received
meaning changed the receiver's behavior beneficially. Measuring it requires
simulation or real testbeds with sequential decision
tasks~\cite{arxiv260113685,arxiv260935936,arxiv260208675}.
\end{enumerate}

\section{Conclusion}
\label{sec:conclusion}
This survey has traced semantic communication from its origins in Weaver's
distinction between technical, semantic, and effectiveness problems, through
the DeepJSCC and DeepSC wave of 2019--2022, the task- and knowledge-driven wave
of 2022--2024, to the generative, foundation-model, and agentic wave of
2024--2026. A unified encoder--channel--decoder model with an explicit shared
knowledge base was used to show how these seemingly disparate threads are
instances of the same design problem: choose \emph{what to transmit} so that a
receiver with \emph{what it already knows} can do \emph{what it needs to do},
at minimum cost. Six comparative tables and five original diagrams synthesize
the theory, coding, modality, reasoning, generative, network, security, and
standardization literature; the 169 verified references, 84.0\% of which are
from 2023 onward, document the field's recent acceleration.

The central unresolved issue is not engineering but \emph{definition}: until
semantic distortion, task utility, and the knowledge base itself admit
operational, comparable definitions, the field will continue to produce
impressive demonstrations that cannot be rigorously ranked. Progress therefore
depends as much on shared benchmarks, coding workflows, and standards as on
larger models. The most promising research direction is the coupling of
foundation models with explicit, verifiable semantic state---tokens, plans, or
knowledge graphs---so that generative power can be audited rather than merely
admired.


\end{document}